\documentclass[sigconf]{acmart} 

\AtBeginDocument{%
  \providecommand\BibTeX{{%
    \normalfont B\kern-0.5em{\scshape i\kern-0.25em b}\kern-0.8em\TeX}}}

\setcopyright{none}
\renewcommand\footnotetextcopyrightpermission[1]{}

\usepackage{array} 
\usepackage{multirow}
\usepackage{slashbox}
\usepackage{algpseudocode}
\usepackage{algorithm, algpseudocode}
\usepackage{amsmath}
\usepackage{calc}
\usepackage{booktabs}

\begin{document}

\title{Self-Supervised Pretraining for Heterogeneous Hypergraph Neural Networks}

\author{Abdalgader Abubaker*}
\email{agader@meta.com}
\thanks{*This work was done when all the authors were at Meta.}
\affiliation{
  \institution{Meta AI}
  \city{London}
  \country{United Kingdom}
}

\author{Takanori Maehara*}
\email{tmaehara@meta.com}
\affiliation{
  \institution{Meta AI}
  \city{London}
  \country{United Kingdom}
}

\author{Madhav Nimishakavi*}
\email{madhavn@meta.com}
\affiliation{
  \institution{Meta AI}
  \city{London}
  \country{United Kingdom}
}

\author{Vassilis Plachouras*}
\email{vplachouras@meta.com}
\affiliation{
  \institution{Meta AI}
  \city{London}
  \country{United Kingdom}
}


\begin{abstract}
Recently, pretraining methods for the Graph Neural Networks (GNNs) have been successful at learning effective representations from unlabeled graph data. However, most of these methods rely on pairwise relations in the graph and do not capture the underling higher-order relations between entities. Hypergraphs are versatile and expressive structures that can effectively model higher-order relationships among entities in the data. Despite the efforts to adapt GNNs to hypergraphs (HyperGNN), there are currently no fully self-supervised pretraining methods for HyperGNN on heterogeneous hypergraphs. 
 
In this paper, we present SPHH, a novel \textbf{s}elf-supervised \textbf{p}retraining framework for \textbf{h}eterogeneous \textbf{H}yperGNNs. Our method is able to effectively capture higher-order relations among entities in the data in a self-supervised manner. SPHH is consist of two self-supervised pretraining tasks that aim to simultaneously learn both local and global representations of the entities in the hypergraph by using informative representations derived from the hypergraph structure. Overall, our work presents a significant advancement in the field of self-supervised pretraining of HyperGNNs, and has the potential to improve the performance of various graph-based downstream tasks such as node classification and link prediction tasks which are mapped to hypergraph configuration. Our experiments on two real-world benchmarks using four different HyperGNN models show that our proposed SPHH framework consistently outperforms state-of-the-art baselines in various downstream tasks. The results demonstrate that SPHH is able to improve the performance of various HyperGNN models in various downstream tasks, regardless of their architecture or complexity, which highlights the robustness of our framework.   
\end{abstract}



\maketitle
\pagestyle{plain}
\section{Introduction}
\label{sec:introduction}

Graph Neural Networks (GNNs) have drawn lots of attention in recent years because of their success in different machine-learning applications that involve graphs~\cite{GNNBook2022, wu2020comprehensive, zhou2022graph}.
GNNs efficiently learn representations of node-attributed graphs, which are graphs whose nodes have feature vectors~\cite{kipf2016semi}; see Section~\ref{sec:overview-on-graph-neural-networks}.

One weakness of traditional GNNs is that they only capture pairwise relations between the nodes.
As many real-world systems involve complex relationships between entities, which are often modelled as higher-order interactions rather than pairwise interactions, the traditional GNNs fail to capture such relationships.
For example, in a web-based recommendation system~\cite{wu2022graph}, a group of people watching a movie together is a higher-order interaction and cannot be represented by pairwise interaction alone.
Such complex relationships can be represented by a \textit{hypergraph}~\cite{bretto2013hypergraph}, an expressive mathematical structure that is able to model higher-order relations and complex interactions. 
Note that this example also motivates us to consider another real-world complexity of the heterogeneity of data, i.e., there are different types of entities and relations between those entities.

Since the hypergraph structure is a powerful tool for representing higher-order relationships among the entities, several works have developed models learning from hypergraphs which largely belong to two categories.
The first category is spectral-based methods, which generalize the spectral methods of ordinary graphs to directly learn from hypergraphs ~\cite{zhou2006learning, agarwal2006high, feng2018learning}.
The second category is the GNN-based methods, which applies GNN after converting hypergraphs to ordinary graphs by expansion methods~\cite{yadati2019hypergcn, bai2021hypergraph, hgnn_plus}.
Our method proposed in this paper is applied to any of these categories. 
We select to use the second category in our work.
We call the models that belong to the second category as \textit{HyperGNN}.


Although many works have addressed the pretraining on GNNs, there are only a few that address the pretraining of HyperGNNs.
In this work, we propose SPHH, a novel self-supervised framework for pretraining HyperGNNs on heterogeneous hypergraphs.
We show the effectiveness of the SPHH pretraining framework on several HyperGNN models over different graph-based downstream tasks that we map to a hypergraph configuration (we simply say downstream tasks).
Our pretraining framework generates embeddings that carry a local representation and a global representation \cite{hu2019strategies}.
SPHH learns the local representation by accumulating information on the neighborhood's attributes, while it learns the global representation by capturing the higher-order relations within the hypergraph.
The pretrained HyperGNN can then be used as initialization of a model used for different  downstream tasks (which we termed as \emph{downstream models}).
All of our experiments show the advantage of the HyperGNN models that are pretrained with our SPHH pretraining framework. 

Our contributions in this paper are the following:
\begin{itemize}
    \item We propose SPHH, a novel, fully self-supervised pretraining framework for HyperGNNs that captures the higher-order relations in \textit{heterogeneous} hypergraphs. The proposed framework can be used to enhance the performance of  several HyperGNN models in different downstream tasks.
    \item We design two  self-supervised pretraining tasks to capture the local and the global representation on the heterogeneous hypergraph. We also define a negative sampling approach over hyperedges. 
    \item We demonstrate the effectiveness of SPHH by conducting several experiments with different HyperGNN models on two public datasets. 
\end{itemize}

The rest of the paper is organized as follows.
Section~\ref{sec:preliminaries} provides the basic definitions and necessary backgrounds that are needed for the rest of the sections.
Section~\ref{sec:proposed_method} introduces our proposed SPHH pretraining framework.
Section~\ref{sec:experiments} presents our experimental result and discuss
them. 
Section~\ref{sec:related_works} discusses related works, and we conclude this paper in Section~\ref{sec:conclusion}.


\section{Preliminaries}
\label{sec:preliminaries}
\subsection{Definitions}
\label{sec:definitions}

\paragraph{\textbf{Graph}}
A \emph{node-attributed graph} (we call it an \emph{ordinary graph}, or simply a \emph{graph} to distinguish from hypergraphs) is a tuple $\mathcal{G} = (\mathcal{V}, \mathcal{E}, \mathcal{X})$ that consists of the set of nodes $\mathcal{V}= \{v_1, v_2, \dots, v_n \}$, the set of edges $\mathcal{E} = \{e_1, \dots, e_m\}$ where each edge $e_i = (u_i, v_i)$ is a pair of nodes, and the attribute matrix $\mathcal{X}$, which is defined on the set of nodes $\mathcal{V}$. 
An edge $e = (u, v) \in \mathcal{E}$ is \emph{undirected} if the reverse edge $(v, u)$ is also in $\mathcal{E}$. 
We denote by $\mathcal{E}_v = \{e \in \mathcal{E} : v \in e \}$ the set of edges incident to $v$.
The \emph{neighborhood} $\mathcal{N}_{\mathcal{G}}(v)$ of node $v$ is defined by the set of all nodes that are connected with $v$, i.e., $\mathcal{N}_{\mathcal{G}}(v) = \bigcup_{e \in \mathcal{F}_v} e \setminus \{ v \}$.
We omit the subscript and denote by $\mathcal{N}(v)$ if the context is clear.
Each node $u$ has a \emph{node type} $t_V(u) \in \{ 1, \dots, T_V \}$, which groups nodes by their properties.
Similarly, each edge $e$ has an \emph{edge type} $t_E(e) \in \{1, \dots, T_E \}$, which groups edges by their properties.
The node types and edge types induce partitions of nodes and edges as 
$\mathcal{V} = \mathcal{V}_1 \cup \dots \cup \mathcal{V}_{T_\mathcal{V}}$ and $\mathcal{E} = \mathcal{E}_1 \cup \dots \cup \mathcal{E}_{T_\mathcal{E}}$, respectively.
$\mathcal{G}$ is \emph{homogeneous} if $T_V = T_E = 1$ and \emph{heterogeneous} otherwise. 
In this paper, we only consider heterogeneous graphs as they are strictly more general than homogeneous graphs.

\paragraph{\textbf{Hypergraph}}
A \emph{node-attributed hypergraph} (we simply call it a \emph{hypergraph}) is a tuple $\mathcal{H} = (\mathcal{V}, \mathcal{F}, \mathcal{X})$ that consists of the set of nodes $\mathcal{V}$ and the set of higher-order relations called \emph{hyperedges} $\mathcal{F} = \{f_1, \dots, f_m\}$, where each hyperedge $f_i$ is a subset of the nodes,  i.e., $f_i \subseteq \mathcal{V}$, and the attribute matrix $\mathcal{X}$ on the set of nodes $\mathcal{V}$.
We denote by $\mathcal{F}_v = \{f \in \mathcal{F} : v \in f \} $ the set of hyperedges incident to $v$.
The \emph{neighborhood} $\mathcal{N}_{\mathcal{H}}(v)$ of node $v$ is defined similarly to the graph case. 
We omit the subscript and denote by $\mathcal{N}(v)$ if the context is clear.
The \emph{heterogeneous} and \emph{homogeneous} hypergraphs are defined similarly to the heterogeneous graph case.
In this paper, we consider heterogeneous hypergraphs with multiple types of nodes and one type of hyperedges.

\paragraph{\textbf{Hypergraph Clique Expansion}}

The \emph{clique expansion}~\cite{zhou2006learning} is a procedure that converts a homogeneous hypergraph $\mathcal{H}$ to an ordinary homogeneous graph by replacing every hyperedge with a clique, i.e., $\mathcal{E}(\mathcal{F}) := \bigcup_{f \in \mathcal{F}} \{(u, v): u, v \in f \}$.
This procedure was used as a preprocessing for learning from hypergraphs~\cite{agarwal2006high, yadati2019hypergcn, feng2019hypergraph}.

Here, we extend the clique expansion to heterogeneous hypergraphs.
This replaces every hyperedge by a clique except edges between the nodes of the same types to reduce the edge intensity and decrease the noise degree, i.e., the resulting ordinary graph has the following set of the edges:
\begin{align}
    \mathcal{E}(\mathcal{F}) := \bigcup_{f \in \mathcal{F}} \{ (u, v) : u, v \in f, t_V(u) \neq t_V(v) \}.
\end{align}

\begin{figure}[tb]
  \centering
  \includegraphics[width=\linewidth]{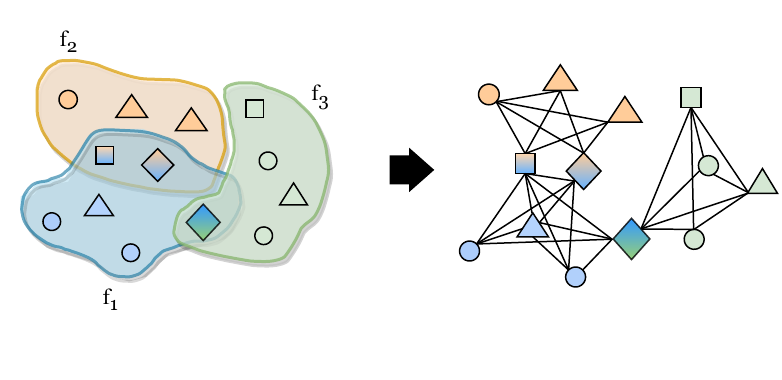}
  \caption{ Clique expansion of a heterogeneous hypergraph with three hyperedges $f_1$, $f_2$, and $f_3$.}
  \label{fig:clique}
\end{figure}

\subsection{Overview on Graph Neural Networks}
\label{sec:overview-on-graph-neural-networks}

GNNs learn a representation for node $v$ by utilizing the attributes of the neighborhoods $N(v)$ by the \emph{neural message passing}~\cite{gilmer2017neural}.
The procedure consists of multiple iterations.
At the $k$-th iteration, each node $v$ updates its own \emph{hidden representation} $h^{(k)}_{v}$ by the following formula:
\begin{equation}
    h^{(k+1)}_{v} = \mathit{Update} \left( h^{(k)}_{v}, \mathit{Aggregate}(\{h^{(k)}_{u}, \forall u \in \mathcal{N}(v)\})\right),
\end{equation}
where $\mathit{Update(.)}$ and $\mathit{Aggregate(.)}$ are arbitrary differentiable functions (e.g., neural networks).
The output of the aggregation function is called \emph{message}.
The aggregation function $\mathit{Aggregate(.)}$ is a permutation invariant function that takes the set of embeddings of the nodes in $\mathcal{N}(v)$ and generates a message.
Then, the update function $\mathit{Update(.)}$ combines the message with the embedding $h^{k}_{v}$ of node $v$ at the $k$-th iteration to generate the updated embedding $h^{(k+1)}_{v}$.
Since $\mathit{Aggregate(.)}$ is a permutation invariant function, the GNNs that are defined in this way are node-permutation equivariant (a.k.a., isomorphism equivariant)~\cite{hamilton2020graph, bronstein2021geometric}.
The specific form of $\mathit{Update(.)}$ and $\mathit{Aggregate(.)}$ determine the architecture of a GNN.
For example, the popular GraphSAGE~\cite{sage} model uses a \emph{mean pooling}, \emph{max pooling}, or \emph{LSTM}~\cite{hochreiter1997long} functions as $\mathit{Aggregate(.)}$.
Then, it concatenates the message with the previous node embedding by a fully-connected neural network (e.g., multi-layer perceptrons MLPs) which forms $\mathit{Update(.)}$.
Thus, the GraphSAGE layer with \emph{mean} aggregator is expressed as follows:
\begin{equation*}
     h^{(k+1)}_{v} = \mathit{MLP} \left(h^{(k)}_{v} \mathbin\Vert  \mathit{mean}(\{h^{(k)}_{u}, \forall u \in \mathcal{N}(v)\}) \right),
\end{equation*}
where $\Vert$ is the concatenation of vectors.

\begin{figure*}[t]
  \includegraphics[width=\textwidth]{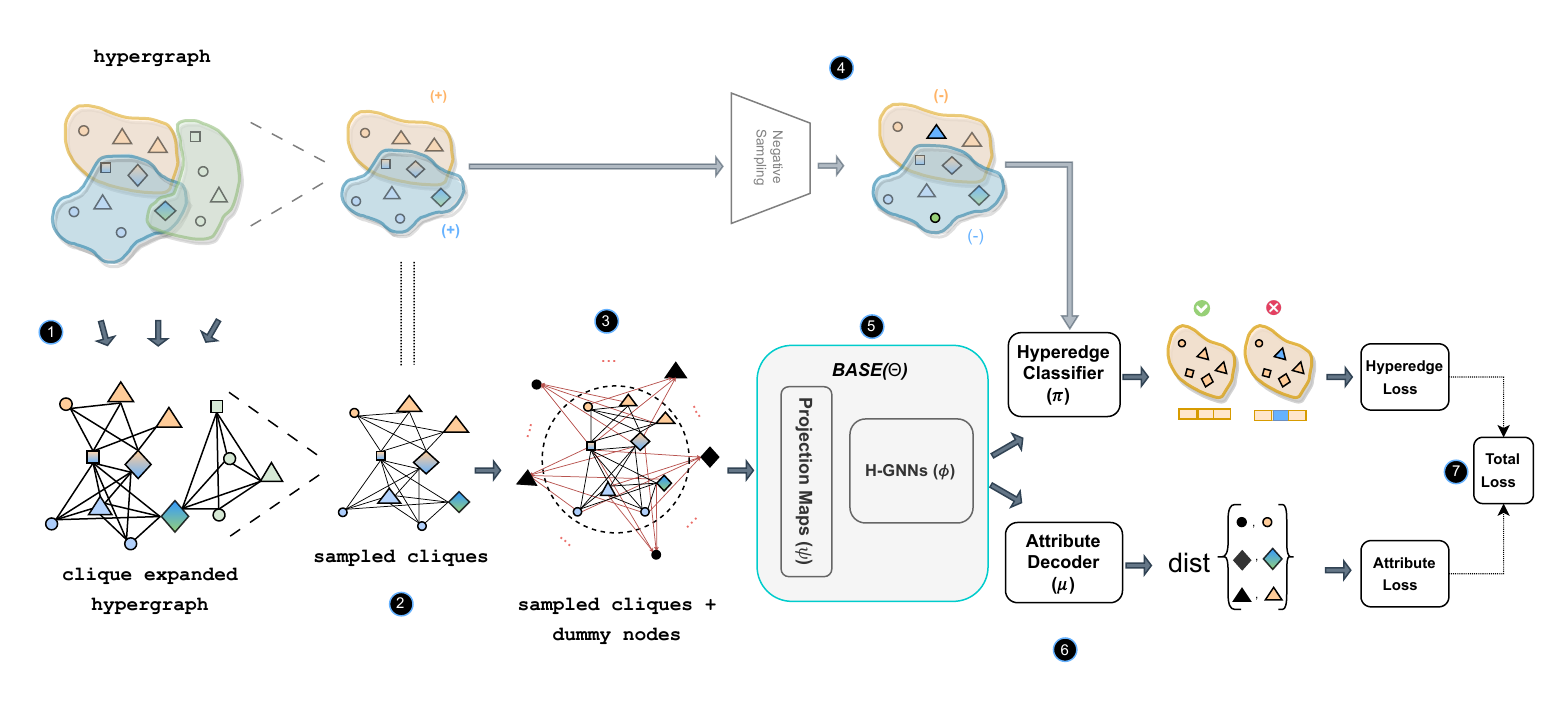}
  \caption{The overall pretraining procedure of our SPHH framework. 1- Expanding the heterogeneous hypergraph with clique expansion. 2- Sample cliques which associated with some positive hyperedges. 3- Adding the dummy nodes with incoming edges from the neighborhoods. 4- Sample negative hyperedges for the positive hyperedges. 5- Feed the overall graph to the $\mathit{BASE}$ model. 6- Perform two pretraining tasks and compute the loss for each task. 7- Calculate the weighted (total) loss. After convergence, we transfer the $\mathit{BASE}$ model for downstream tasks.} 
  \label{fig:full}
\end{figure*}

\subsection{Pretraining Graph Neural Networks}

Model pretraining has recently demonstrated incredible success in different machine-learning applications in several areas such as Natural Language Processing~\cite{devlin2018bert, reimers2019sentence, brown2020language, zhang2022opt} and Computer Vision~\cite{he2020momentum, chen2020simple}.
Generally, pretraining is a highly effective technique for enhancing the performance of neural networks on different tasks by training the model on a large data set to learn robust and generalizable features and then finetuning on a smaller and task-specific dataset.
Pretraining has two categories:
1) Unsupervised or self-supervised pretraining that trains a neural network to discover underlying patterns in the input data without labeled data.
2) Supervised pretraining that trains a neural network to predict tasks that are related to the output task using labeled data.   

This success has extended to GNNs, where pretraining is used to improve the performance of the GNNs on different graph-related tasks by learning general and robust node representations using large unlabeled graph-structured data~\cite{hu2020gpt, hu2019strategies, yang2022self, qiu2020gcc}.
The GNNs utilize pretraining to capture the intrinsic structure of the graph and the node attribute patterns underlying the graph.
However, the problem of pretraining HyperGNNs is not well-explored.
In our work, we consider the transfer learning settings, where we initially pretrain a generic HyperGNN with our SPHH pretraining framework using specific self-supervised pretraining tasks.
Then, we use the pretrained HyperGNN as model initialization  to improve model performance on different downstream tasks.






\section{Proposed Method}
\label{sec:proposed_method}

In this section we introduce \emph{SPHH}, a self-supervised pretraining framework for HyperGNNs,
which is inspired by Hu~\textit{et al.}~\cite{hu2019strategies} on strategies for pretraining GNNs.
SPHH consists of two pretraining tasks to learn useful local and global representations simultaneously.
In Section~\ref{sec:model}, we demonstrate the HyperGNN model that we used as an encoder.
In Section~\ref{sec:nac}, we present how our model learns the local representation through the task of \emph{Node Attribute Construction}.
Section~\ref{sec:hp} illustrates the other pretraining task \emph{Hyperedge Prediction}, where the model learns the global representation given the higher-order relations within the hypergraph.
The overall approach is presented in Figure~\ref{fig:full}.

\subsection{Model}
\label{sec:model}
As we stated in the Section~\ref{sec:introduction}, there are two categories of methods to learn on hypergraphs.
Firstly, methods that generalize the spectral mechanisms on ordinary graphs to directly learn from hypergraphs to generate rich node representations.
The other category is the methods that adapt the GNNs to hypergraph by firstly expanding the hypergraph to ordinary graph using any expansion technique, e.g., clique expansion, then use standard GNNs on the expanded hypergraph to learn the node embeddings.
Here we called the methods that belong to this category by HyperGNNs.
Our SPHH pretraining framework could be applied to both categories.
In our paper, we select the second category in which we use the clique expansion followed by standard GNN as an encoder.

Since we consider the heterogeneous hypergraph that has multiple types of nodes and one type of hyperedges, we use type-specific learnable projection maps to  project different types of node attributes into the same dimensional space. Moreover, we convert the homogeneous GNN model to a heterogeneous GNN (\textit{H-GNN}) model by implement the message and update functions individually for each edge type in the (expanded) heterogeneous graph $\mathcal{G}$.

Overall, the model we pretrain, which we called $\mathit{BASE}$,  consists of two parts: a type-specific \emph{learnable projections} $P_{t_V}(\psi)$ and the \emph{heterogeneous graph neural network} H-GNN($\phi$). So, the $\mathit{BASE}$ model (encoder) takes the clique-expanded heterogeneous hypergraph as input and generates the node embedding for each node type (see Figure~\ref{fig:full}). In the following two sections, we demonstrate the two self-supervised pertaining tasks of SPHH framework. The architecture of the $BASE$ model is used by existing works to adjust the standard GNNs to heterogeneous data \cite{wang2019heterogeneous, hu2020heterogeneous}.   

\subsection{Node Attribute Construction}
\label{sec:nac}

\begin{figure}[h]
  \centering
  \includegraphics[width=\linewidth]{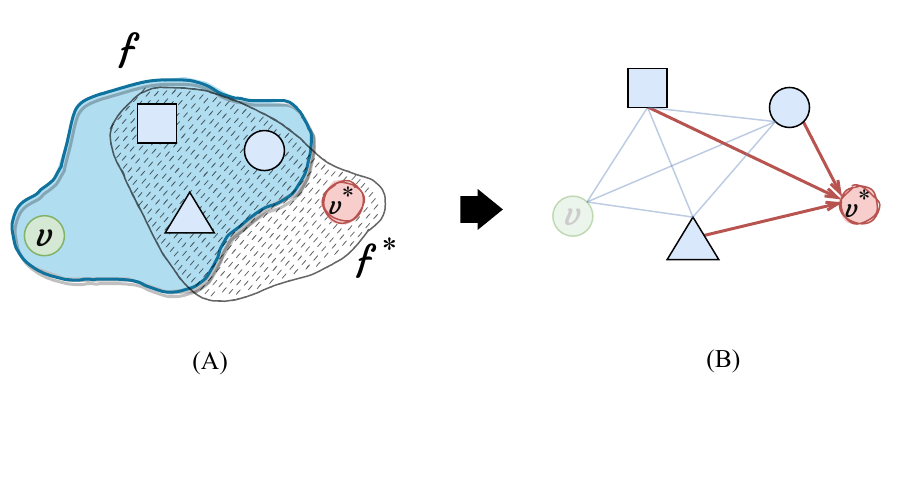}
  \caption{ (A) Neighborhood sharing between the node $v$ and its associated dummy node $v^*$. (B) The dummy node $v^*$ aggregate  the information from the neighborhood .}
  \label{fig:neigh_share}
\end{figure}

Inspired by \cite{hu2020gpt}, the goal of node attribute construction task $\mathcal{T}_c$ is to make the model learn how to \textit{construct} node attributes of a specific node given the attributes in its neighborhood.
The motivation behind this task is to learn local representation by generating node embeddings, whereby the nodes inside the hyperedge are close in the embedding space.
Given a hypergraph $\mathcal{H} = (\mathcal{V}, \mathcal{F}, \mathcal{X})$, for every node $v \in \mathcal{V}$ and its local neighborhood $\mathcal{N}(v)$, we attempt to maximize the likelihood $p( x_v|\mathcal{X}_{\mathcal{N}(v)}, \mathcal{F}_v )$.
To train this task, we create a \textit{dummy} node $v^*$ for every node $v \in \mathcal{V}$; they are only used in this task.
The dummy node $v^*$ has a random attribute $x_{v^*}$, which has the same dimensionality as $x_v$. The attributes assigned to dummy nodes of the same type are identical, and they are calculated by taking the average of the attributes of all nodes of that type.
Then, for every hyperedge $f \in \mathcal{F}_v$ incident to $v$, we create the associated hyperedge $f^* = f \setminus \{ v \} \cup \{ v^* \}$ that is incident to the dummy node $v^*$ (see Figure~\ref{fig:neigh_share}~(A)). For the purpose of the message passing operation, this is implemented as follows:  for each dummy node $v^*$ we construct \emph{incoming} edges from every node that connected with node $v$ to the dummy node $v^*$. The purpose of creating the dummy nodes is to prevent any information leaking during message passing (see Figure~\ref{fig:neigh_share}~(B)).
Then, the hidden representation of the dummy node $v^*$ at iteration $k$ is computed as follows:
\begin{equation}\label{eq:nac}
    h_{v^*}^{(k)} = \mathit{BASE}\left(\Theta^{k-1}, \mathcal{G}^{*},\mathcal{X}_{\{ v^* \} \cup \mathcal{N}(v)}\right),
\end{equation}
where $\mathit{BASE}$ is the architecture introduced in Section~\ref{sec:model} and
$\Theta^{k-1}$ represents the set of base model's parameters $\{ \psi, \phi \}$. $\mathcal{G}^{*}$ is an auxiliary graph created by adding a dummy node $v^*$ connected to the neighborhood $\mathcal{N}(v)$ through incoming edges for all in $v \in \mathcal{V}$.

Thus, the attribute construction loss is given by
\begin{equation}\label{eq:nc_loss}
    \mathcal{L}_{\mathcal{T}_c} = \frac{1}{|\mathcal{V}|} \sum_{v, v^* \in \mathcal{V}} \mathit{dist}(x_{v}, \mathit{DEC}(h_{v^*})),
\end{equation}
where $\mathit{dist}$ is any distance function, and $\mathit{DEC}$ is a type-specific fully connected neural network to decode an embedding to the input feature. 
By training the model by this task, it learns to construct any node attributes from its neighborhood.
Thereby, the model will learn a useful local representation through the node attribute construction task $\mathcal{T}_c$.

\subsection{Hyperedge Prediction}
\label{sec:hp}

Concurrently with the node attribute construction task $\mathcal{T}_c$, we perform the task of hyperedge prediction $\mathcal{T}_p$.
The goal of this task is to learn a useful global representation by learning to discriminate between the ground-truth higher-order relations and \emph{perturbed} higher-order relations.
The ground-truth higher-order relations are the members of $\mathcal{F}$.
They are called \emph{positive hyperedges}, and are denoted by $f^+$.
The perturbed higher-order relations are generated by the \emph{negative sampling} method that modifies every hyperedge $f^+$ by randomly replacing a subset of its nodes with nodes from $\mathcal{V} \setminus f^+$; see below for the detailed description.
The perturbed hyperedges are called \emph{negative hyperedges} and are denoted by $f^-$.
We denote the set of all positive and negative hyperedges by $\mathcal{F}^{(+,-)}$.
Then, the task $\mathcal{T}_p$ is a hyperedge classification problem that distinguishes positive and negative edges in $\mathcal{F}^{(+,-)}$.


\paragraph{Negative Sampling} 

Given a heterogeneous hypergraph $\mathcal{H} = (\mathcal{V}, \mathcal{F},\mathcal{X})$, we define the negative sampling  over the family of hyperedges $\mathcal{F}$ as follows.
For every positive hyperedge $f^+$ and for a randomly-selected node type $t$, we randomly select an $\alpha$-fraction subset $S_t$ from $f^+ \cap \mathcal{V}_t$, which is the nodes of type $t$.
Then, we select a random subset $R_t \subseteq \mathcal{V}_{t} \setminus f^+$ with $|R_t| = |S_t|$. 
The corresponding negative hyperedge is then defined by $f^- = f^+ \setminus S_t  \cup R_t $.

\begin{figure}[ht]
  \centering
  \includegraphics[width=\linewidth]{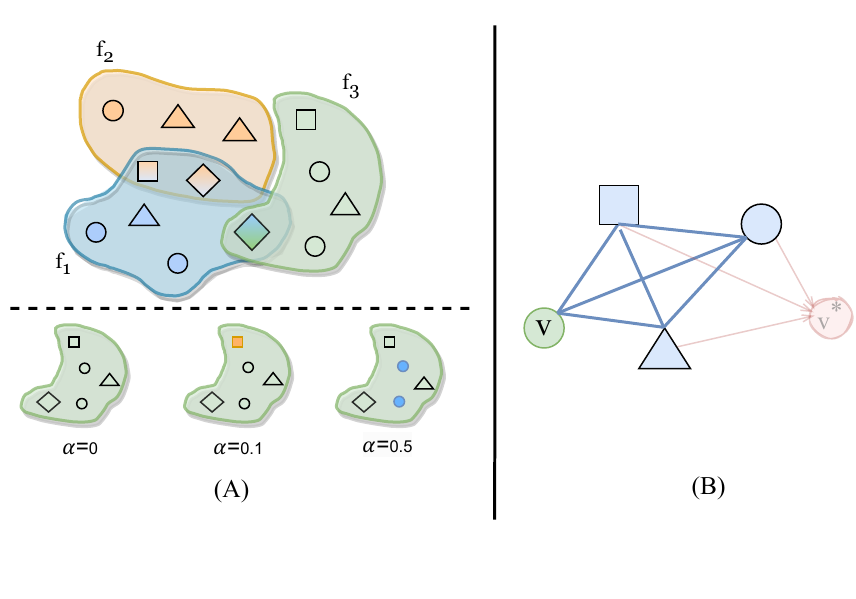}
  \caption{ (A) An example showing three negative hyperedges as a result of perturbation with $\alpha = \{0, 0.2, 0.5\}$ on the ground-truth hyperedge $f_3$. (B) Normal message passing between node $v\in \mathcal{V}$ and its neighborhood only.}
  \label{fig:hp_ngs}
\end{figure}

The perturbed ratio $\alpha$ is a hyperparameter used to control the amount of change applied to the positive hyperedge. 
It is inversely proportional to the similarity between the positive and the negative hyperedges.
Thus, if $\alpha$ took small value then it will be harder for the model to discriminate between the positive and corresponding negative hyperedges, as both hyperedges tend to be similar. Figure~\ref{fig:hp_ngs}~(A) below demonstrates the negative sampling by showing an example of a perturbed positive hyperedge for three different values of $\alpha$.
\newline

The hidden node representations are then used by the hyperedge prediction task $\mathcal{T}_p$ are computed using the \textit{primary} expanded heterogeneous hypergraph $\mathcal{G}$ (excluding the dummy nodes), in which the message passing appears between the actual nodes and their neighbors (see Figure~\ref{fig:hp_ngs}~(B)).
Accordingly, the hidden representation of node $v$ at iteration k is computed as follows:

\begin{equation}
    h_{v}^{(k)} = \mathit{BASE}(\Theta^{k-1}, \mathcal{G},\mathcal{X}_{\{ v \} \cup \mathcal{N}(v)}).
\end{equation}
Here, the model parameters $\Theta^{k-1}$ are shared with the  Equation~\eqref{eq:nac} for the node attribute generation task.

Now, given the hidden node representations, we define the \textit{hyperedge representation $\eta_f$} for a hyperedge $f \in \mathcal{F}^{(+,-)}$ as an aggregation of all hidden representations for every node inside the hyperedge $f$, i.e.  
\begin{equation}\label{eq:hyper_rep}
    \eta_f = \mathit{Aggregate}(\{h_{v} : v \in f\}),
\end{equation}
where the $\mathit{Aggregate}$ is any permutation invariant function. In the forward pass, we compute the hyperedge representations for all positive and negative hyperedges.
Then we pass these representations to binary classifier $\tau_\pi$, parameterised by $\pi$, to predict whether these hyperedge representations are positive or negative. 
Thus, the loss of the hyperedge prediction task $\mathcal{T}_p$ is computed via binary cross entropy:

\begin{align}\label{eq:hp_loss}
    \mathcal{L}_{\mathcal{T}_p} = 
    \frac{1}{\lambda} \sum_{f \in \mathcal{F}^{(+,-)}} 
    \left( \frac{}{} \right. & y_f \cdot \log \sigma(\tau_{\pi}(\eta_f)) \nonumber \\[-10pt]
    & + \left. (1 - y_f) \cdot \log (1 - \sigma(\tau_{\pi}(\eta_f))) \right),
\end{align}
where $\lambda = |\mathcal{F}^{(+,-)}|$, $\sigma$ is sigmoid function, and $y_f$ is the class of the hyperedge $f$. 

\hspace{2cm}

\subsection*{Total Loss}

Finally, we calculate the overall loss $\mathcal{L}$ as a weighted sum of aforementioned pretraining tasks losses in the equations \eqref{eq:nac} and \eqref{eq:hp_loss}:
\begin{equation}\label{eq:loss}
    \mathcal{L} = \beta \mathcal{L}_{\mathcal{T}_c} + (1-\beta)\mathcal{L}_{\mathcal{T}_p}, 
\end{equation}
where $\beta \in [0, 1]$ is a hyperparameter.

Having the loss, we then use the gradient descent to optimize all the learnable parameters.
Once the parameters are optimized, we use the pretrained weights $\Theta$ of the $\mathit{BASE}$ for the downstream model initialization. Algorithm~\ref{alg:cap} presents the pseudocode for a mini-batch pretraining procedure of the SPHH framework. In the following section, we report the results from  experiments, which show that many downstream models benefit from SPHH pretraining strategy in several downstream tasks.

\begin{algorithm}
\caption{SPHH Pretraining algorithm}\label{alg:cap}
\begin{algorithmic}[1]
\Require  \noindent\begin{itemize}
    \item[-] Attributed heterogeneous hypergraph $\mathcal{H}= (\mathcal{V}, \mathcal{F}, \mathcal{X})$,
    \item[-] The clique expansion method \textit{clique\_expansion(.)},  
    \item[-] Clique Sampler \textit{Sampler(.)}, 
    \item[-] To heterogeneous GNN converter \textit{to\_hetero(.)}
\end{itemize}

\Ensure 
\State Expand $\mathcal{H}$ by  $\mathcal{G} = clique\_expansion(\mathcal{H})$
\State Initialize the $\mathit{BASE (\Theta)}$ model which consists of projection maps $P_{t_V}(\psi)$ and the heterogeneous GNN H-GNN$(\phi)$).
\State For each node type $t$, calculate the dummy node attributes $r_t$ by averaging all node attributes $X_t$.  

\For{each sampled cliques $G \in \mathit{Sampler}(.)$}
    \For{each node $v_t$ from type $t$ in $G$}
        \State \parbox[t]{\linewidth-3em}{Create a dummy node $v_{t}^*$ with a random attribute $x_{v_{t}^*} = r_t$.}
        \State \parbox[t]{\linewidth-3em}{Connect $v_{t}^*$ using incoming edges from every node that is linked with node $v$ to formulate the overall subgraph $G\cup G^* $.}
    \EndFor
    
    \State \parbox[t]{\linewidth-1.5em}{Apply the $\mathit{BASE (\Theta)}$ to the overall subgraph $G\cup G^* $ and compute the node embeddings.}
    \State \parbox[t]{\linewidth-1.5em}{Divide the node embeddings into two groups: dummy node embeddings and original node embeddings.}
    \State \parbox[t]{\linewidth-1.5em}{Calculate the loss $\mathcal{T}_c$ in Equation~\eqref{eq:nc_loss} using the dummy node embeddings and the loss $\mathcal{T}_p$ in Equation~\eqref{eq:hp_loss} using the original node embeddings.}
    \State \parbox[t]{\linewidth-1.5em}{Calculate the total weighted loss $\mathcal{L}$ as in Equation~\eqref{eq:loss}.}
    \State \parbox[t]{\linewidth-1.5em}{Optimize $\Theta$ by minimizing the loss $\mathcal{L}$.}
\EndFor
\State{\Return The pretrained $\mathit{BASE(\Theta^*})$ for downstream tasks use.}
\end{algorithmic}
\end{algorithm}


\section{Experiments}
\label{sec:experiments}
We conduct several experiments to evaluate our proposed SPHH pretraining strategy.
We evaluate our strategy based on the downstream model's performance on various graph tasks.
Specifically, we are interested on the performance gain when the downstream model weights are initialized by the weights that are pretrained with our SPHH pretraining strategy. 

We evaluate the efficiency of our pretraining strategy by transferring the pretrained node embeddings into two graph downstream tasks: \emph{node classification task} and \emph{link-prediction} task. 
Both downstream tasks are mapped into hypergraph configuration (more details in Section~\ref{sec:fine-tuning}). For both downstream tasks, we fine-tune with an assumption of limited access to the labeled data during the training, similar to real-world applications when it is infeasible to access a sufficient amount of labeled data. 

\subsection{Datasets}
 
\textbf{OGB-MAG(H)}: The public OGB-MAG~\cite{mag} is a widely used benchmark heterogeneous academic graph dataset, and it is a subset of Microsoft Academic Graph (MAG).
The dataset includes four types of nodes: \textit{papers} (736,389), \textit{authors} (1,134,649), \textit{institutions} (8,740), and \textit{fields of study} (59,965).
Each node has a 128-dimensional attribute vector.
The paper node attributes are the average of the \textit{word2vec} embeddings~\cite{word2vec} of words in its title and abstract using.
The attribute vectors for all other node types are computed using \textit{metapath2vec} \cite{dong2017metapath2vec}.
All papers are also associated with the year of publication, which ranges from 2010 to 2019.

We create the hypergraph version \textit{OGB-MAG(H)} by connecting the nodes through the hyperedges as follows. 
Every hyperedge connects a paper with all authors who contribute on the paper, all institutions associated with each individual author, and the paper's fields of study.
Accordingly, the number of hyperedges $|\mathcal{F}|$ is equal to the number of papers in the hypergraph.
The statistics of the hypergraph version \textit{OGB-MAG(H)}  are highlighted in Table~\ref{table:datasets}. 

The downstream tasks that we define on \textit{OGB-MAG(H)} are: 1) \textit{node classification} to  predict the paper's venue from 349 different classes~\cite{mag}, and 2) \textit{link prediction}, where given a paper and a candidate author, we predict whether the  candidate author has contributed to the paper~\cite{hu2020gpt}.

\noindent \textbf{AF-Reviews(H)}: Amazon Fashion Reviews dataset is a category of Amazon Review Data~\cite{amazon}.
It consists of three types of nodes: \textit{reviews} (882,403), \textit{users} (748,219) , and \textit{fashion products} (186,112).
The reviews were collected for users who made a review on a fashion product between 2002 to 2018.
Each user has at most one review for each product. 
Each review has an associated rating from 1 to 5.
Each node has associated with a 384-dimensional attribute vector.
Firstly, we use the pretrained  \textit{Sentence-BERT} model~\cite{reimers2019sentence} to generate an embedding vector of every review using the associated texts, where the generated embedding is used as review's attribute vector.
Thereafter, following the same approach in \cite{hu2020gpt}, we compute the attributes of a user by averaging the attributes of all reviews that have been written by the user.
Likewise, the attributes of the product node type is computed by averaging the attributes of all reviews that have been written for the product.  

We construct the hypergraph version \textit{AF-Reviews(H)} in which a hyperedge connects each review with the user who made the review and the corresponding fashion product.
Since each review is associated with only single user and single fashion product, then each hyperedge has exactly three nodes.
Given that the hyperedge is connect every review with its associated user and product, then the number of hyperedges $|\mathcal{F}|$ is equal to the number of reviews in the hypergraph. 
Table~\ref{table:datasets} also presents the statistics of \textit{AF-Reviews(H)} hypergraph.

The downstream task that we define on \textit{AF-Reviews(H)} is \textit{node classification}, to predict the review's rating from 5 different rate classes~\cite{hu2020gpt}. 


\begin{table}[tb]
  \caption{Summary of the datasets. $\max(|f_i|)$ and $\min(|f_i|)$ denote the maximum and minimum numbers of nodes in the hyperedges, respectively.}
  \label{table:datasets}
    \centering
    \begin{tabular}{ccc}
            \toprule
        &  \textbf{OGB-MAG(H)} & \textbf{AF-Reviews(H)} \\
        \midrule
            {\# node types}    &       4           &              3                \\
            {$|\mathcal{V}|$ }   &   1,939,743       &            1,816,734          \\
            {$|\mathcal{F}|$}     &   736,389         &         882,403                \\
            {$\max(|f_i|)$}       &    5,623           &           3                   \\
             {$\min(|f_i|)$}      &       2           &           3                   \\
    \bottomrule
    \end{tabular}
\end{table}

\subsection{Experimental Setup}

\subsubsection{Dataset splitting} \hfill\\
The goal of pretraining is to transfer knowledge learned from many unlabeled data in order to facilitate the downstream tasks when there are only few labels. In our benchmarks, we pretrain and fine-tune using data from different time spans. We split each dataset as follows: \textit{pretraining/prevalidation} sets for pretraining stage, and \textit{train/validation/test} sets for fine-tuning stage.

In \textit{OGB-MAG(H)}, all papers published before the end of 2015 are used as pretraining set and all papers published in 2016 are used as prevalidation set. 
During fine-tuning, all papers published between 2017 and 2018 are used for the train set and papers published in 2018 and since 2019 are used for valid and test sets, respectively.

In \textit{AF-Review(H)}, all fashion product reviews posted before  2013  are used as pretraining set, where the reviews posted in 2014 are used as prevalidation set.
During fine-tuning, all reviews posted between 2015 and 2017 are used for the train set and reviews posted in 2018 and in 2019 are used for valid and test sets, respectively.

\subsubsection{Pretraining\label{pretraining}} \hfill\\
In this section, we demonstrate the whole procedure and the setup of the SPHH pretraining.
As data preprocessing step, we firstly convert the heterogeneous hypergraph $\mathcal{H}$ to the standard heterogeneous graph $\mathcal{G}$ using the clique expansion method. Then we generate the mini-batches of the heterogeneous graph $\mathcal{G}$. The sampled subgraph can be seen as set of cliques. For example, in the expanded \textit{OGB-MAG(H)}, a clique is a set consisting of one paper, all its authors, all related institutions, and all related fields of study that are fully-connected using different edge types and  no edges between the nodes of the same types. When we sample a subgraph, we construct the auxiliary subgraph that created by adding a dummy node $v^*$ for every node $v$ in the subgraph, and connect $v^*$ to all $u \in \mathcal{N}(v)$ using incoming edges. The final graph will be fed as input to the $\mathit{BASE}$ model. To generate the negative hyperedges, we use the negative sampling method with perturbed ratio $\alpha = 0.1$ to generate one negative hyperedge instance for every positive hyperedge associated with its respective clique in the subgraph.

The encoder $\mathit{BASE}$ is defined as follows: the first part of the base model is learnable projections $P_{t_V}(\psi)$.
We use number of multi-layer perceptrons (MLPs) equal to the number of node types to map all attributes of the different node types (including the dummy nodes) to the same dimensional space.
The second part is the heterogeneous graph neural network H-GNN$(\phi)$.
Due to the fact that the standard message passing GNNs cannot be applied directly to the heterogeneous data, as the heterogeneous graph data can not be processed by the same functions due to the differences in the feature types in the nodes and the edges.
We use a build-in function called `\verb|to_hetero()|'' provided in the PyTorch Geometric (PyG) library to convert the standard GNNs to heterogeneous GNNs by implement the message and update functions individually for each edge type in $\mathcal{G}$.
So, we use the heterogeneous GNN to produce the node embedding for the input heterogeneous graph.
In our experiments, we evaluate our pretraining strategy using four standard GNNs: \textit{GraphSAGE} \cite{sage}, \textit{Graph Isomorphism Networks} (GIN) \cite{gin}, \textit{Graph Attention Networks} (GAT) \cite{gat}, and \textit{GraphConv} \cite{graphconv}.
We divide the node embeddings into two groups: the embeddings of dummy nodes that used for task of node attribute construction $\mathcal{T}_c$ (Equation~\eqref{eq:nc_loss}), and the embeddings of original nodes that are used to compute the hyperedge representation $\eta$ in the task of the hyperedge prediction $\mathcal{T}_p$ (Equation~\eqref{eq:hyper_rep}).
Then, we compute the weighted loss (Equation~\eqref{eq:loss}) of the two task and then optimize all learnable parameters using the gradient descent.

Finally, we use the prevalidation data to evaluate the performance of the model in the two \textit{pretraining} tasks.
After the convergence, we transfer the pretrained parameters of the $\mathit{BASE}$ model for downstream model initialization in the fine-tuning stage.

\subsubsection{Fine-Tuning}\hfill\\
\label{sec:fine-tuning}
In the fine-tuning stage, we attach a linear classifier to the pretrained $\mathit{BASE}$ model for both node classification task and link prediction task.
Then, we fine-tune the pretrained model with only $5\%$ and $15\%$ of labeled data during training to meet the assumption that we have limited access to labeled data as in real-life applications.
To leverage the higher-order relations, which are represented by hyperedges, in the final embeddings computations, we mapped the node classification task and link-prediction task into the hyperedge configuration. 
In the task of node classification, to predict the class of a node $u \in \mathcal{V}$, we compute the final embedding $z_u$ by aggregating the nodes embeddings of all nodes that sharing a hyperedge with $u$ which can be expressed as follows:
\begin{equation*}
    z_u = \mathit{Aggregate} \left(\{ v \in f : f \in \mathcal{F}_u \} \right).
\end{equation*}
Then, $z_u$ entered as input to the linear classifier to get the classes scores. In the task of link prediction, we model the problem as a hyperedge prediction task by learning to predict the existence of an edge between $u$ and $v$ if the edge's end-nodes $u$ and $v$ are appeared in the same hyperedge. In other words, the model is learning to discriminate between the positive hyperedge that include the edge's end-nodes and the negative hyperedge that perturbed one of the edge's end node only. For example, the link prediction task on \textit{OGB-MAG(H)} to predict that a specific author of the paper is mapped as follow: given a positive hyperedge that contain a paper, all authors, all institutions, and the fields of study, we generate the negative hyperedges by perturbing only the specific author by replacing the author with other author outside the hyperedge. Similar to the hyperedge prediction pretraining task $\mathcal{T}_p$, we model the problem as a binary classification problem where the only difference is in the negative sampling. Rather than computing the score of the dot product of the edge end-node embeddings, we use a linear classifier to compute the score for the positive and negative \textit{edges} by feeding the hyperedge embeddings $\eta$ of the corresponding hyperedges.

\subsubsection{Baseline models}\hfill\\
To show the gain of pretraining with SPHH, as baselines, we compare the following HyperGNN models when they randomly initialized against the same models when they pretrained with SPHH in the task of node classification and link prediction as they defined in the previous section.
\begin{itemize}
    \item \textit{GraphSAGE}, by Hamilton~\textit{et al.}\cite{sage}.
    \item \textit{Graph Isomorphism Network (GIN)}, by Xu ~\textit{et al.}~\cite{gin}.
    \item \textit{Graph Attention Network (GAT)}, by Veli{\v{c}}kovi{\'c}~\textit{et al.}~\cite{velivckovic2018deep}. 
    \item \textit{GraphConv}, by Morris~\textit{et al.}~\cite{graphconv}.
\end{itemize}

\subsection{Experiments Settings and Hyperparameters}
  Our implementation is built on top of PyG library. We use Tesla P100 SXM2 16GB for all pretraining and fine-tuning experiments. We use  random search to optimize the set of hyperparameters during pretraining and fine-tuning for all GNN models. In pretraining, the set of hyperparameters during pretraining involve learning rate, batch size, dropout ratio, hidden dimensions,  aggregation function, perturbed ratio $\alpha$, and loss weight $\beta$. We pretrain each experiment for $100$ epochs. Additionally, we search the hyperparameters during fine-tuning for both experiments with $5\%$ and $15\%$ and we fine-tune each experiments for $100$ epochs. For optimization, we use the Adam~\cite{kingma2014adam} algorithm to optimize the loss function. Table~\ref{table:pretrain-hyper} show the hyperparameters we used in our experiments.

\begin{table}[tb]
  \caption{Hyperparameters used in our experiments for both pretraining and fine-tuning. The asterisk (*) indicates that a  hyperparameter is only used for pretraining.}
  \label{table:pretrain-hyper}
    \centering
    \begin{tabular}{cc}
            \toprule
        Hyperparameters &  Values \\
        \midrule
            Number of layers              &    2, 3             \\
            Hidden dimensions                &    (128, 256),  (256, 512)             \\
            Learning rate                   &       $10^{-3}$, $10^{-4}$, $10^{-5}$              \\
            Batch size                      &   128, 256, 512, 1024, 2048         \\
            Dropout                         &   0.2, 0.3, 0.4, 0.5, 0.6       \\
            Aggregation function                     &   max, mean, sum            \\
             Perturbed ratio ($\alpha$)$^*$      &       $0.1, 0.2$            \\
            Loss weight ($\beta$)$^*$             &    $0.6, 0.55$             \\
           
    \bottomrule
    \end{tabular}
\end{table}



\subsection{Results}
In this section, we report the experimental results of different HyperGNN models pretrained with our proposed SPHH pretraining framework on \textit{OGB-MAG(H)} and \textit{AF-Reviews(H)} heterogeneous hypergraph benchmarks. In almost all experiments, all HyperGNN models that are pretrained with SPHH  outperform the non-pretrained models in various graph downstream tasks.

\subsubsection{Node classification}\hfill\\
Table~\ref{table:node-classification} shows the experimental results on the defined node classification task for both \textit{OGB-MAG(H)} and \textit{AF-Reviews(H)} datasets. The table is divided into two blocks vertically. The upper block present the performance of different HyperGNN models when we fine-tuning with $5\%$ of the labeled data, while the lower block show the models performance when we fine-tuning with $15\%$ of the labeled. We report \textit{Accuracy} ($\%$) and  \textit{F1-Macro} for the test set. In all experiments, the accuracy of pretrained HyperGNN models are outperforming the baselines (non-pretrained HyperGNN models). Similarly, almost all results on the F1-Macro metric, except two with (*), show the superiority of pretrained HyperGNN model over the baselines. The experiments of GAT on \textit{OGB-MAG(H)} ran out of memory (OOM in the table) during the  computation of the node representation for batches that include hyperedges with very large number of nodes, this due to attention mechanism that learn multiple edge weights, which does not allow use sparse-matrix multiplication in the GAT implementation in PyG library.

\begin{table*}
\caption{Results of Node classification task for \textit{OGB-MAG(H)} and \textit{AF-Reviews(H)} datasets. We report the Accuracy and F1-Macro metrics for experiments of pretrained and non-pretrained different HyperGNNs models when using $5\%$ and $15\%$ of labeled data. OOM indicates an out-of-memory error, and the asterisk (*) indicates the non-pretrained outperform the pretrained model.}
\label{table:node-classification}
\begin{tabular}{rrcccc}
\toprule 
& & \multicolumn{2}{c}{\textbf{OBG-MAG(H)}} & \multicolumn{2}{c}{\textbf{AF-Reviews(H)}} \\
& & Accuracy & F1-Macro & Accuracy & F1-Macro \\
\midrule
\textbf{5\% labeled data} \\
\quad \multirow[t]{2}{*}{GraphSAGE} & non-pretrained & $20.77 \pm 0.77$ & $0.0512 \pm 0.0008$ & $62.92 \pm 0.09$ & $0.3473 \pm 0.002$ \\
                                    & pretrained & $\mathbf{22.25 \pm 0.72}$ & $\mathbf{0.0563 \pm 0.004}$ & $\mathbf{63.55 \pm 0.13}$ & $\mathbf{0.3604 \pm 0.004}$ \\[2pt]
\quad \multirow[t]{2}{*}{GIN} & non-pretrained & $18.22 \pm 0.83$ & $0.0168 \pm 0.0001$ & $61.70 \pm 0.26$ & $0.2915 \pm 0.007$ \\
                              & pretrained & $\mathbf{18.30 \pm 2.0}$ & $\mathbf{0.0370 \pm 0.001}$ & $\mathbf{61.74 \pm 0.45}$ & $\mathbf{0.2993 \pm 0.005}$ \\[2pt]
\quad \multirow[t]{2}{*}{GAT} & non-pretrained & OOM & OOM & $61.10 \pm 0.31$ & $0.2964 \pm 0.006$ \\
                              & pretrained & OOM & OOM & $\mathbf{63.89 \pm 0.08}$ & $\mathbf{0.3519 \pm 0.002}$ \\[2pt]
\quad \multirow[t]{2}{*}{GraphConv} & non-pretrained & $20.90 \pm 0.54$ & $0.0554 \pm 0.002$ & $63.13 \pm 0.11$ & $0.3527 \pm 0.001$ \\
                                    & pretrained & $\mathbf{23.23 \pm 0.14}$ & $\mathbf{0.0632 \pm 0.001}$ & $\mathbf{63.60 \pm 0.29}$ & $\mathbf{0.3607 \pm 0.002}$ \\[2pt]
\textbf{15\% labeled data} \\
\quad \multirow[t]{2}{*}{GraphSAGE} & non-pretrained & $28.09 \pm 0.2$ & $0.0913 \pm 0.0006$ & $64.73 \pm 0.10$ & $\mathbf{0.3851 \pm 0.003^{*}}$ \\
                                    & pretrained & $\mathbf{29.5 \pm 0.5}$ & $\mathbf{0.0997 \pm 0.0004}$ & $\mathbf{64.95 \pm 0.10}$ & $0.3820 \pm 0.007$ \\[2pt]
\quad \multirow[t]{2}{*}{GIN} & non-pretrained & $20.59 \pm 0.31$ & $0.0349 \pm 0.01$ & $61.00 \pm 2.07$ & $0.3213 \pm 0.0118$ \\
                              & pretrained & $\mathbf{24.18 \pm 2.17}$ & $\mathbf{0.0671 \pm 0.001}$ & $\mathbf{62.50 \pm 0.83}$ & $\mathbf{0.3417 \pm 0.0092}$ \\[2pt]
\quad \multirow[t]{2}{*}{GAT} & non-pretrained & OOM & OOM & $63.64 \pm 0.18$ & $0.3538 \pm 0.003$ \\
                              & pretrained & OOM & OOM & $\mathbf{65.03 \pm 0.08}$ & $\mathbf{0.3763 \pm 0.002}$ \\[2pt]
\quad \multirow[t]{2}{*}{GraphConv} & non-pretrained & $29.21 \pm 0.61$ & $0.0972 \pm 0.001$ & $64.77 \pm 0.10$ & $\mathbf{0.3852 \pm 0.003^{*}}$ \\
                                    & pretrained & $\mathbf{30.12 \pm 0.41}$ & $\mathbf{0.1051 \pm 0.0006}$ & $\mathbf{65.09 \pm 0.10}$ & $0.3788 \pm 0.001$ \\
\bottomrule
\end{tabular}
\end{table*}

\subsubsection{Link prediction}\hfill\\
We also evaluate the effectiveness of SPHH over the several HyperGNN models on the task of link prediction that defined on the\textit{ OGB-MAG(H)} dataset. We meseaure the performance using the \textit{AUC} metric. Similarly, we evaluate the model performance when we fine-tune using $5\%$ and $15\%$ of the labeled data. Table~\ref{table:link-prediction} present the results of different HyperGNN models that pretrained with SPHH versus same models that  randomly initialized. The results show that all HyperGNN models are benefit from the SPHH pretraining and they outperform the non-pretrained models.   

\begin{table}
\caption{Results of link prediction task on \textit{OGB-MAG(H)} dataset. We report AUC for experiments of pretrained and non-pretrained GNNs models, when using $5\%$ and $15\%$ of labeled data. OOM indicates an out-of-memory error.}
\label{table:link-prediction}
\begin{tabular}{rrcc}
\toprule 
& & \multicolumn{1}{c}{\textbf{OBG-MAG(H)}} \\
& & AUC \\
\midrule
\textbf{5\% labeled data} \\
\quad \multirow[t]{2}{*}{GraphSAGE} & non-pretrained & $0.9643 \pm 0.0016$ \\
                                    & pretrained & $\mathbf{0.9799 \pm 0.0015}$ \\[2pt]
\quad \multirow[t]{2}{*}{GIN} & non-pretrained & $0.7300 \pm 0.0160$ \\
                              & pretrained & $\mathbf{0.9001 \pm 0.0290}$ \\[2pt]
\quad \multirow[t]{2}{*}{GAT} & non-pretrained & OOM \\
                              & pretrained & OOM \\
\quad \multirow[t]{2}{*}{GraphConv} & non-pretrained & $0.9666 \pm 0.0017$ \\
                                    & pretrained & $\mathbf{0.9854 \pm 0.0019}$ \\[2pt]
\textbf{15\% labeled data} \\
\quad \multirow[t]{2}{*}{GraphSAGE} & non-pretrained & $0.9797 \pm 0.0037$ \\
                                    & pretrained & $\mathbf{0.9845 \pm 0.0005}$ \\[2pt]
\quad \multirow[t]{2}{*}{GIN} & non-pretrained & $0.9194 \pm 0.0170$  \\
                              & pretrained & $\mathbf{0.9568 \pm 0.0070}$\\[2pt]
\quad \multirow[t]{2}{*}{GAT} & non-pretrained & OOM \\
                              & pretrained & OOM \\[2pt]
\quad \multirow[t]{2}{*}{GraphConv} & non-pretrained & $0.9816 \pm 0.0004$ \\
                                    & pretrained & $\mathbf{0.9886 \pm 0.0006}$ \\
\bottomrule
\end{tabular}
\end{table}

\subsubsection{Results Discussion}\hfill\\
In Table~\ref{table:node-classification} of the node classification results, we can see a performance gain in the accuracy due to pretraining with SPHH framework in all the HyperGNN models. The gain in the accuracy is shown in both OGB-MAG(H) and AF-Reviews(H) benchmarks and for the two training label ratios $5\%$ and $15\%$.  The accuracy gain due to SPHH pertaining is become higher when we fine-tune with few label data in most cases. If we compare the accuracy gain between using $5\%$ and $15\%$ of labeled data in fine-tuning,  the accuracy gains of most HyperGNN models like in (GraphSAGE, GraphConv, and GAT) due to pretraining with SPHH framework when we fine-tuning with only $5\%$ of labeled data is higher than the accuracy gain when we fine-tune using $15\%$ of labeled data. 

Likewise, in the link prediction task, as we can see in Table~\ref{table:link-prediction}, the performance gain in the AUC due to the pretraining with SPHH framework is become higher when we fine-tune using only $5\%$ of OGB-MAG(H) labeled data is higher with large margin against the AUC gain when fine-tune using $15\%$ of the labeled data.

Generally, in most cases, the efficiency of pretraining with SPHH framework is greater when we have less labeled data in the downstream tasks, similar to the cases in real-life applications.


\section{Related Work}
\label{sec:related_works}
In this section we review the related work on learning with hypergraphs, and pretraining GNNs on ordinary graphs and hypergraphs.

\subsection{Learning on Hypergraphs}
 The concept of using machine-learning on hypergraph and the clique expansion was firstly introduced in a ground-breaking study by Zhou~\textit{et al.}~\cite{zhou2006learning} to generalize 
the powerful methodology of spectral clustering that operate on undirected graphs to hypergraphs, and further develop algorithms for hypergraph embedding and transductive inference. Then the clique expansion become widely-used in~\cite{agarwal2006high, satchidanand2015extended, feng2018learning, yadati2019hypergcn}. In recent years, several works adapting the GNNs to hypergraph because its expressiveness. Feng~\textit{et al.}~\cite{feng2018learning} is the first to generalize the convolution operation to the hypergraph learning process by introducing the hypergraph neural networks ($\textit{HGNN}$) which has been extended recently to ($\mathit{HGNN}^{+}$) in~\cite{hgnn_plus}. HyperGCN~\cite{yadati2019hypergcn} proposed a semi-supervised method for training the graph convolution network (GCN) on hypergraph using spectral theory of hypergraph. Bai~\textit{et al.}~\cite{bai2021hypergraph} generalize of the GCN and GAT on hypergraphs by proposing two end-to-end trainable operators: hypergraph convolution and hypergraph attention. DHE~\cite{payne2019deep} introduce a framework that combines the use of context and permutation-invariant vertex membership properties of hyperedges in hypergraphs to carry out both transductive and inductive learning for classification and regression. Huang~\textit{et al.}~\cite{huang2021unignn} has proposed the UniGNN, a unified framework to interpret the message passing in graphs and hypergraphs. All previous methods on hypergraphs learn in a supervised manner. Our proposed approach is fully self-supervised as it does not need any labeled data.
\subsection{Pretraining Graph Neural Network with Graph}
Several works has proposed unsupervised learning methods on graphs. Kipf~\textit{et .al}~\cite{kipf2016semi} introduced the  variational graph autoencoder (VGAE) a framework for unsupervised learning on graph-structured data which try to reconstruct the graph's adjacency matrix and learn interpretable latent representations.  Also, Hamilton~\textit{et al.}\cite{sage} design an unsupervised loss function to train the GraphSAGE without task-specific supervision, and Veli{\v{c}}kovi{\'c}~\textit{et al.}~\cite{velivckovic2018deep}
introduce the deep graph infomax (DGI), a unsupervised learning method that operate on maximise the mutual information between node representations.
Recently, several works has addressed pretraining GNN on graph-structured data and transfer the pretrained model for downstream tasks.
Hu~\textit{et al.} develop effective pretraining strategy for GNN, and show that combining the node-level and graph-level task enhance the performance on graph classification task.
Also, GPT-GNN,  which is introduced by Hu~\textit{et al.}~\cite{hu2020gpt}, provide a generative pretraining framework for GNN models by learning to reconstruct the attribute and the structure of the input graph.
GCC in~\cite{qiu2020gcc} use contrastive learning approach to pretraining the GNNs.
All previous pretraining strategies are not capturing the higher-order relations but rely on pairwise relations.
Our pretraining framework is capturing the higher-order relations through operating on hypergraphs. 

\subsection{Pretraining Hypergraph Neural Networks}
The utilization of hypergraphs for pretraining HyperGNNs is limited.
To the best of our knowledge there is only one recent work conducting this track  which is proposed by Du~\textit{et al.}~\cite{du2021hypergraph}.
The proposed method called \textit{HyperGene}, a pretraining strategy for HyperGNNs using homogeneous hypergraph which is incorporates two self-supervised pretraining tasks.
They try to capture the local context through the first node-level task by learn to distinguish between the node inside and outside the hyperedge where the node context is all other nodes inside the same hyperedge.
The second task is aim to learn hyperedge similarity in which similar hyperedges are tend to be in same cluster, the task is modelled as hyperedge clustering.


Our work is different from them as we consider the heterogeneous case, which is a more practical setting. 
Both we and they pretrain a model by node-level task and hyperedge-level task, but both tasks are different as follows.
Our node-level task is a node attribute generation, which generates node attributes using neighbors' attributes,
whereas theirs is a classification task to decide whether a node is a member of a hyperedge.
Our hyperedge-level task is a hyperedge classification task to discriminate between positive and negative hyperedges, 
whereas theirs is a clustering problem by predicting the membership of the hyperedge to a cluster.




\section{Conclusion \& Future Works}
\label{sec:conclusion}
We have introduced SPHH, a novel self-supervised pretraining framework for HyperGNNs that operates on heterogeneous hypergraphs. Our method captures higher-order relationships among entities in a self-supervised manner. SPHH is composed of two self-supervised tasks that aim to learn both local and global representations of entities.  The local node representation is learned  by generate the node attribute given the the attributes of all nodes inside each hyperedge that contain the node of interest. The global representation is learned by distinguish between the between positive and negative hyperedges. Our experiments on two real-world benchmarks using four HyperGNN models show that SPHH consistently outperforms state-of-the-art baselines and improves performance in various graph-based tasks, demonstrating its robustness.

In the future, we plan to generalize the SPHH framework by considering multiple hyperedge types, and to improve effectiveness with attention mechanisms.




\clearpage 
\bibliographystyle{ACM-Reference-Format}
\bibliography{ref}

\end{document}